\documentclass{article} 
\usepackage{iclr2027_conference,times}

\usepackage{amsmath,amsfonts,bm}

\def\eqref#1{equation~\ref{#1}}

\def\1{\bm{1}}

\DeclareMathAlphabet{\mathsfit}{\encodingdefault}{\sfdefault}{m}{sl}
\SetMathAlphabet{\mathsfit}{bold}{\encodingdefault}{\sfdefault}{bx}{n}

\newcommand{\R}{\mathbb{R}}

\DeclareMathOperator{\reshape}{reshape}
\DeclareMathOperator{\permute}{permute}

\usepackage[british]{babel}
\usepackage{hyperref}
\usepackage{url}
\usepackage{graphicx}
\usepackage{adjustbox}
\usepackage{booktabs}
\usepackage{array}
\usepackage{tabularx}
\usepackage{multirow}
\usepackage{enumitem}
\usepackage[T1]{fontenc}
\usepackage[varqu,varl]{zi4}
\usepackage{listings}
\newcommand{\code}{\lstinline}

\title{DanLing NestedTensor: Composable Multi-Ragged Tensors for Deep Learning}

\author{Zhiyuan Chen\thanks{\url{zyc.ai}}\\
  DanLing Team\\
  \texttt{this@zyc.ai}
}

\iclrfinalcopy 
\begin{document}

\maketitle

\lhead{Preprint}
\vspace{-3.6em}
\providecommand{\NTArtifactRoot}{.}
\providecommand{\NTMechanismSpeedMin}{1.79}
\providecommand{\NTMechanismSpeedMax}{2.26}
\providecommand{\NTBertEagerSpeed}{3.88}
\providecommand{\NTBertCompiledSpeed}{4.32}
\providecommand{\NTBertMemorySaving}{61.95}
\providecommand{\NTFCNEagerSpeed}{2.08}
\providecommand{\NTIMDBEagerGM}{2.74}
\providecommand{\NTWikiTextEagerGM}{1.32}
\providecommand{\NTIMDBCompiledGM}{3.39}
\providecommand{\NTADEEagerGM}{1.97}
\providecommand{\NTWMTEagerRatio}{0.67}
\providecommand{\NTWMTTraceSpeed}{58.18}
\providecommand{\NTWMTSteadySpeed}{1.60}
\providecommand{\NTWMTRecompiles}{four}
\providecommand{\NTPairPadMin}{2.40}
\providecommand{\NTPairPadMax}{4.32}
\providecommand{\NTPairPackedMin}{1.50}
\providecommand{\NTPairPackedMax}{3.94}
\providecommand{\NTPairMemPad}{38.08}
\providecommand{\NTPairMemNested}{5.41}
\providecommand{\NTPairCompiledMin}{1.06}
\providecommand{\NTPairCompiledMax}{2.20}
\providecommand{\NTPairCompiledRecompiles}{two}
\providecommand{\NTPairFusionOffMsMin}{522}
\providecommand{\NTPairFusionOffMsMax}{556}
\providecommand{\NTPairFusionOffEagerMin}{1.13}
\providecommand{\NTPairFusionOffEagerMax}{1.17}
\providecommand{\NTPairFusionOffDanLingMin}{2.23}
\providecommand{\NTPairFusionOffDanLingMax}{6.08}
\providecommand{\NTPairIfaceCostMax}{1.60}
\providecommand{\NTPairGranMin}{1.53}
\providecommand{\NTPairGranMax}{3.94}
\providecommand{\NTPairMemPadMin}{34.76}
\providecommand{\NTPairMemPadMax}{38.09}
\providecommand{\NTPairUniformLengths}{208--224}
\providecommand{\NTPairUniformOcc}{94.4}
\providecommand{\NTFCNMemPad}{28.38}
\providecommand{\NTFCNMemNested}{12.25}
\providecommand{\NTViTEagerGM}{0.91}
\providecommand{\NTViTMemGM}{1.48}
\providecommand{\NTDETRSpeed}{1.05}
\providecommand{\NTDETRMemEff}{4.27}
\providecommand{\NTMechanismOverheadMin}{25}
\providecommand{\NTMechanismOverheadMax}{34}
\providecommand{\NTCtlAttnEager}{1.47}
\providecommand{\NTCtlAttnCompiled}{1.51}
\providecommand{\NTCtlDanLingEager}{3.94}
\providecommand{\NTCtlDanLingCompiled}{4.31}
\providecommand{\NTCtlBucketEager}{1.01}
\providecommand{\NTCtlBucketCompiled}{1.17}
\providecommand{\NTCtlCompiledGap}{0.19}
\providecommand{\NTIMDBOccRandom}{31.0}
\providecommand{\NTIMDBOccBucket}{97.7}
\providecommand{\NTRetainedCells}{six}
\providecommand{\NTKeptWikiText}{63}
\providecommand{\NTTotalWikiText}{64}
\providecommand{\NTKeptWMTBase}{124}
\providecommand{\NTTotalWMTBase}{128}
\providecommand{\NTKeptWMTBig}{125}
\providecommand{\NTTotalWMTBig}{128}
\providecommand{\NTKeptWikiTextXL}{four}
\providecommand{\NTTotalWikiTextXL}{six}
\providecommand{\NTRetainShiftXL}{4.1}
\providecommand{\NTRetainShiftOther}{0.48}
\providecommand{\NTCompileReuseCells}{five}
\providecommand{\NTCompileCells}{six}
\providecommand{\NTXLSteps}{six}
\providecommand{\NTXLRecompileMin}{once}
\providecommand{\NTXLRecompileMax}{twice}
\providecommand{\NTOperatorCount}{404}
\providecommand{\NTOperatorEntries}{708}

\begin{abstract}
  Variable-size inputs are common in deep learning, but dense batching allocates a shared envelope and spends computation on padding.
  The cost multiplies across varying axes: an explicit pair state allocates $BN_{\max}^2$ positions instead of $\sum_iN_i^2$.
  Packing removes that waste, but composing packed operations still requires the logical axes and sample boundaries a flat buffer no longer exposes.
  We present DanLing NestedTensor, a PyTorch tensor abstraction that makes multi-ragged structure a property of the tensor itself.
  Packed values carry tensor-backed partitions and logical dimension order, so broadcasting creates ragged axes, feature transformations retain them, and reductions consume them.
  The same representation carries through autograd and both eager and compiled execution.
  On an A100, the geometric-mean speedup over same-mode padding is \NTIMDBEagerGM{}$\times$ eager and \NTIMDBCompiledGM{}$\times$ compiled across four BERT scales, and \NTADEEagerGM{}$\times$ eager across four FCN backbones.
  A four-block Pairformer-style workload runs \NTPairPadMin{}--\NTPairPadMax{}$\times$ faster than a padded reference using native PyTorch kernels across square length regimes in eager execution, with peak allocation falling from \NTPairMemPad{} to \NTPairMemNested{}~GiB on its high-variation batch.
  The tensor interface lets model code built from its supported operators compose efficient variable-size computation without managing offsets at any call site.
  \ificlrfinal
  Code is available at \url{https://github.com/ZhiyuanChen/DanLing}.
  \else
  Code will be released publicly upon publication.
  \fi
\end{abstract}

\vspace{-1em}
\begin{figure}[htbp]
  \centering
  \includegraphics[width=\linewidth]{\NTArtifactRoot/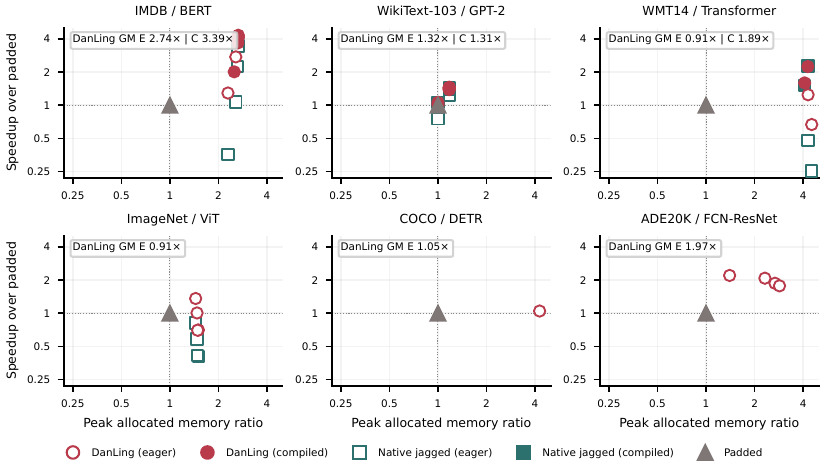}
  \caption{\textbf{Speed and memory efficiency relative to same-mode padded execution.}
    One panel per workload; one point per model scale and mode, against that cell's own padded baseline at $(1,1)$.
    Shape and colour identify the method, fill the mode (hollow eager, solid compiled).
    Both axes use steady-state execution (Section~\ref{sec:setup}), so the coordinates share one measurement window; Appendix Table~\ref{tab:compilation} reports compilation counts and full-trace cost.
    Inline text gives DanLing's geometric mean per mode.
  All completed paired measurements are shown; other outcomes use the codes of Appendix~\ref{app:protocol}.}
  \label{fig:headline}
\end{figure}

\section{Introduction}
\label{sec:introduction}
\vspace{-1em}

Data vary in size; dense tensor batches do not.
Sequence packing already handles the one-dimensional case, and for language models it is close to sufficient.
Much of what people train is not one-dimensional: a protein structure model refines an $N_i\times N_i$ residue-pair state through dozens of blocks~\citep{jumper2021alphafold,abramson2024alphafold3}, and vision models are trained at variable resolution and aspect ratio~\citep{dehghani2023navit}.
Convolution must keep height and width separate to address neighbourhoods within each image.
Padding places these examples inside a common envelope, consuming memory and computation at empty positions~\citep{fegade2022cora}.
The cost multiplies across varying axes: a pair representation over a length-$N_i$ sample holds $N_i^2$ entries, so a padded batch allocates $BN_{\max}^2$ of them rather than $\sum_iN_i^2$; a sample at half the maximum length wastes three quarters of its block rather than half.
On our Pairformer-style workload padded allocation barely moves: \NTPairMemPadMin{}--\NTPairMemPadMax{}~GiB whether lengths are near-uniform or badly skewed (Table~\ref{tab:pairformer}).

Collecting that saving is harder than allocating a smaller buffer, because the pair state is not where the program ends.
A single--pair--single block broadcasts two single representations into a pair tensor, a residual feature transformation updates it, and a reduction over one pair axis returns singles.
Packing the valid pair cells removes the padded storage, but the flat buffer no longer shows which cells share a sample and a row.
The reduction needs that grouping, and its result must carry a structure the next operator can accept.
When this structure is maintained outside the tensor, model code must supply indexing and reconstruction logic at each structural transition.

Existing approaches address different parts of this problem.
Bucketing groups examples of similar size, which reduces padding but ties batch composition to input geometry and can change the order and statistics of optimisation~\citep{morishita2017minibatch,kocmi2017curriculum}.
Packing keeps the chosen batch, and variable-length attention kernels execute attention on its packed storage~\citep{dao2022flashattention,dao2024flashattention2}.
Neither specifies how logical axes and sample boundaries are maintained when a model constructs, transforms and reduces an explicit pair state, and PyTorch's \code{torch.jagged} admits only one ragged dimension~\citep{pytorchNested}.

DanLing NestedTensor treats this bookkeeping the way ordinary tensors treat shape and dtype: as state the tensor carries, not state the model rebuilds.
The packed payload travels with exact element sizes, hierarchical partitions and logical dimension order, and each handler updates that structure as it computes, so autograd and compilation consume the same representation.
Figure~\ref{fig:headline} summarises the speed and memory this yields; composition is the sharper test.
A four-block Pairformer-style application, which creates, transforms and reduces two ragged pair axes in every block, compiles and trains under this representation with default compiler settings, while the padded and hand-packed implementations we tested do not---although the simpler block compiles for all three.

This paper makes three contributions.
\begin{itemize}[topsep=2pt,itemsep=3pt,parsep=0pt,leftmargin=1.3em]
  \item \textbf{A representation that composes.} Multiple varying logical axes are carried together, including non-leading axes and rectangular pair states, without conflating them with the flattened storage axis. Each handler consults and updates those axes as it already consults shape---broadcasting creates one, feature transformations preserve it, reductions consume it---and returns a structure the next operator can use, so a complete single--pair--single program composes without offset bookkeeping at any call site. Handlers of this kind cover \NTOperatorCount{} distinct operators across PyTorch's two dispatch protocols (Sections~\ref{sec:representation} and~\ref{sec:dispatch}).
  \item \textbf{Differentiation and compilation across the representation boundary.} Bridges in both directions keep the gradient connected where a structured wrapper meets its packed payload, so a compiled block can return singles and pairs to an eager loss and still receive gradients through changes of rank and layout. Compilation fixes the structural schema at each program point while sizes and partitions stay runtime tensor inputs, so supported length-dynamic programs reuse one compiled graph rather than specialising against each length vector (Section~\ref{sec:training-integration}, Table~\ref{tab:compilation}).
  \item \textbf{Efficiency, and where it comes from.} Four BERT scales give a \NTIMDBCompiledGM{}$\times$ compiled geometric-mean speedup over padding and FCN-ResNet50 drops from \NTFCNMemPad{} to \NTFCNMemNested{}~GiB; a four-block Pairformer-style workload runs \NTPairPadMax{}$\times$ faster than a padded reference using native PyTorch kernels on its high-variation batch. Against explicit packing, a matched-granularity control separates cost from strategy: at equal call granularity the tensor interface adds at most \NTPairIfaceCostMax{}\% latency, while batch-level segmented execution runs \NTPairGranMin{}--\NTPairGranMax{}$\times$ faster than per-sample calls (Sections~\ref{sec:workloads} and~\ref{sec:analysis}).
\end{itemize}

\section{Related Work}
\label{sec:related-work}
\vspace{-1em}

\paragraph{Ragged representations.}
TensorFlow RaggedTensor, Awkward Array and FBGEMM jagged operators represent variable-size data through numerical buffers and structural metadata~\citep{tensorflowRagged,pivarski2020awkward,fbgemmJagged}.
PyTorch's native \code{torch.jagged} combines a tensor interface with packed storage, differentiation and compilation for its supported operations; its documented layout admits one ragged dimension~\citep{pytorchNested}.

\paragraph{Ragged and dynamic compilation.}
CoRa provides a programming interface and compiler for ragged computation, using techniques including selective padding and operation splitting~\citep{fegade2022cora}.
PyTorch~2 integrates graph capture, differentiation and backend compilation, while Relax makes symbolic shapes explicit across computational graphs and tensor programs~\citep{ansel2024pytorch,lai2025relax}.
NestedTensor integrates its representation with PyTorch, exposing changing sizes and partitions as tensor data while preserving the structural schema of each program point.

\paragraph{Packing and specialised kernels.}
Sequence packing and NaViT retain independent-example computation while reducing padding in language and vision workloads~\citep{krell2021packing,dehghani2023navit}.
FlashAttention and FlashAttention-2 optimise attention execution, and FlexAttention provides a programmable fused attention interface~\citep{dao2022flashattention,dao2024flashattention2,dong2025flexattention}.
NestedTensor reuses these execution paths while keeping projections, normalisation and residual operations packed around them.
KeOps evaluates formula-defined pairwise computations lazily~\citep{charlier2021keops}, whereas our pair states persist across multiple feature transformations before a structural reduction.

\section{NestedTensor: Representation and Execution}
\label{sec:method}
\vspace{-1em}

\begin{figure}[t]
  \centering
  \includegraphics[width=\linewidth]{\NTArtifactRoot/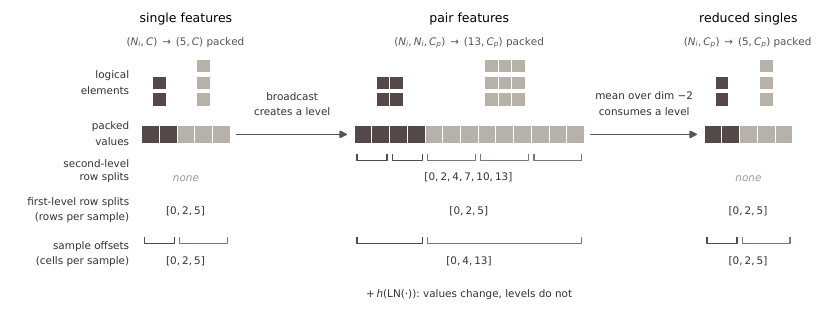}
  \caption{\textbf{Structure through a single--pair--single block.}
    For lengths $(2,3)$, broadcasting expands five single positions into thirteen pair cells; column reduction returns five singles.
    First-level row splits count rows per sample, second-level splits delimit each pair row's cells, and sample offsets delimit complete elements in packed-cell space.
  First-level row splits and sample offsets coincide for singles but differ for pairs.}
  \label{fig:representation}
\end{figure}

A single--pair--single block illustrates how NestedTensor carries logical structure through successive operations.
For nonempty inputs $X_i\in\R^{N_i\times C}$, the block computes
\begin{equation}
  \begin{aligned}
    P_i[a,b,:]
    &= u_\theta(X_i[a,:])+v_\theta(X_i[b,:]),\\
    \widetilde P_i
    &= P_i+h_\theta\!\left(\operatorname{LN}_{C_p}(P_i)\right),\\
    \widetilde X_i[a,:]
    &= X_i[a,:]+g_\theta\!\left(
      \frac{1}{N_i}\sum_{b=0}^{N_i-1}\widetilde P_i[a,b,:]
    \right).
  \end{aligned}
  \label{eq:running-block}
\end{equation}
Here $u_\theta,v_\theta$ project to pair width $C_p$, $h_\theta$ is a SwiGLU~\citep{shazeer2020glu} feature update, and $g_\theta$ maps the reduced features back to width $C$.
The block returns both updated singles and pairs, so the pair state is model state that later operations consume, not a score matrix that exists only inside a fused attention call.
Figure~\ref{fig:representation} draws what the tensor holds at each stage of the block for a two-sample batch, with the exact packed shapes and offsets, and Figure~\ref{fig:program} gives the measured implementation (Appendix~\ref{app:running-program}).

\subsection{Packed Representation of Variable-Size Batches}
\label{sec:representation}

\paragraph{Problem setting.}
A batch $\mathcal X=(X_0,\ldots,X_{B-1})$ contains individually dense tensors of common rank $D$; raggedness is variation between samples, not irregularity inside one.
Let $n_{i,d}$ be the extent of element dimension $d$ in sample $i$.
An ordered tuple $\mathcal R=(r_1,\ldots,r_q)$ declares the ragged axes.
The complementary static axes $\mathcal S=(s_1,\ldots,s_{D-q})$ have common extents $c_j=n_{i,s_j}$ across samples.

\paragraph{Numerical storage.}
The permutation $\pi=(\mathcal R,\mathcal S)$ places ragged axes before static axes.
We collapse each sample's ragged prefix and concatenate the resulting rows.
Writing $p_i=\prod_{j=1}^{q}n_{i,r_j}$ for a sample's packed row count, $T=\sum_i p_i$ and $o_i=\sum_{k<i}p_k$, the value tensor $V\in\R^{T\times c_1\times\cdots\times c_{D-q}}$ satisfies
\begin{equation}
  V[o_i:o_{i+1}]
  =
  \reshape\!\left(
    \permute_{\pi}(X_i),
    (p_i,c_1,\ldots,c_{D-q})
  \right).
  \label{eq:packing}
\end{equation}
Images $(C,H_i,W_i)$ pack as $(\sum_iH_iW_i,C)$, pairs $(N_i,M_i,C)$ as $(\sum_iN_iM_i,C)$, and non-leading layouts $(S,N_i,C)$ as $(\sum_iN_i,S,C)$.

\paragraph{Structural metadata.}
The wrapper retains the exact size matrix $A_{i,d}=n_{i,d}$, sample offsets $o$, dimension order $\pi$, and row splits at each ragged level.
Layouts with explicitly declared ragged axes retain the partitions as tensors, so later operations and reconstructed wrappers can use them directly.
Although sizes can generate these partitions, retaining them avoids recovering per-sample Python shapes at each reconstruction boundary, and exact sizes distinguish zero-volume shapes such as $(0,3)$ and $(0,7)$ whose offsets coincide.
The public logical shape describes the padded envelope; numerical storage remains $V$.

\begin{figure}[t]
\begin{lstlisting}
pair   = self.left_projection(left).unsqueeze(-2) + self.right_projection(right).unsqueeze(-3)
pair   = pair + self.transition(pair)
single = left + self.reduction_projection(pair.mean(dim=-2))

# the same pair construction, packed by hand:
pair   = self.left_projection(left)[rows] + self.right_projection(right)[columns]
\end{lstlisting}
  \caption{\textbf{Nothing in the model code refers to the packing.}
    The measured implementation of Equation~\ref{eq:running-block}, with its three method bodies inlined: \code{left_projection} and \code{right_projection} are $u_\theta$ and $v_\theta$, \code{transition} is the pre-normalised update $h_\theta\circ\operatorname{LN}_{C_p}$, and \code{reduction_projection} is $g_\theta$.
    The tensor carries the partitions of Figure~\ref{fig:representation} through all three lines without appearing in any of them.
  The last line is the explicit-packed baseline's equivalent of the first, reaching the same values but requiring caller-supplied \code{rows} and \code{columns} gather indices.}
  \label{fig:program}
\end{figure}

\subsection{Structure-Aware Operator Dispatch}
\label{sec:dispatch}

A handler interprets its operands in logical coordinates, runs the numerical operation in packed coordinates, and constructs the structure of its result, so that result is ready for the next operator without model-level reconstruction.
NestedTensor installs these handlers on high-level PyTorch functions through \path{__torch_function__} and on ATen operations through \path{__torch_dispatch__}, covering \NTOperatorCount{} distinct operators across the two protocols (Appendix~\ref{app:implementation}).
For ragged coordinates $(a_1,\ldots,a_q)$ in sample $i$, the packed row is
\begin{equation}
  \ell_i(a_1,\ldots,a_q)
  =
  o_i+\sum_{j=1}^{q}a_j\prod_{k=j+1}^{q}n_{i,r_k}.
  \label{eq:coordinate}
\end{equation}
Static axes map individually to axes of $V$, while the ragged coordinates jointly determine its leading index.

\paragraph{Creating axes through broadcasting.}
The projected singles in Equation~\ref{eq:running-block} acquire complementary singleton dimensions, giving element shapes $(N_i,1,C_p)$ and $(1,N_i,C_p)$.
Their broadcast creates $(N_i,N_i,C_p)$ pair states.
The handler associates each pair cell with its sample-local row and column, gathers the corresponding projected features, and constructs the square output sizes and partitions.
Packed row counts change from $N_i$ to $N_i^2$, and in Figure~\ref{fig:representation} the sample offsets from $[0,2,5]$ to $[0,4,13]$.
The first pair partition retains the single-row grouping, while the second records the columns within each row.

\paragraph{Retaining structure through feature transformations.}
Activations and feature-wise normalisation retain the pair axes and partitions.
A projection updates the static feature extent, while the residual addition requires matching logical pair structure.
These operations reuse the existing partitions instead of inferring new ones from their current extents.

\paragraph{Consuming one logical axis.}
For a rectangular pair state $P_i\in\R^{N_i\times M_i\times C_p}$, a column sum $Y_i$ groups cells with the same sample and row coordinate:
\begin{equation}
  V_Y[\eta_i+a,:]
  =
  \sum_{b=0}^{M_i-1}V_P[o_i+aM_i+b,:],
  \qquad
  \eta_i=\sum_{k<i}N_k.
  \label{eq:pair-reduction}
\end{equation}
The output has shape $(N_i,C_p)$ in each sample, with offsets $\eta$ and the remaining row axis.
The square-pair mean in Equation~\ref{eq:running-block} additionally divides each group by $N_i$.
Its projection can then be added to $X_i$ because the handler has restored the single representation's logical structure as well as its values.

\paragraph{Compatibility and semantics.}
Elementwise alignment checks logical extents and dimension order, including the grouping represented by partitions.
For example, shapes $(2,6,C)$ and $(3,4,C)$ both flatten to twelve feature rows but are incompatible for direct elementwise addition.
Dense operands are aligned to logical axes before permutation or gathering.
Each handler is specified so that interpreting its packed output logically agrees with running the reference operation on the logically interpreted inputs, which for sample-local operations is the corresponding dense element computation; Appendix~\ref{app:semantics} states that contract and the semantics of batch and global operations.

\subsection{Efficient Execution on Packed Storage}
\label{sec:execution}

\paragraph{Dense work across valid positions.}
Feature-wise operations reuse dense kernels on $V$.
Projections compute $V'=VW^{\mathsf T}+b$, and normalisation acts on each row's static feature tail.
This preserves a large numerical batch while eliminating padded positions: the feature transformations in Equation~\ref{eq:running-block} process $\sum_iN_i^2$ pair cells rather than $BN_{\max}^2$.

\paragraph{Partition-dependent computation.}
Ragged-axis reductions derive group indices from retained coordinates and use indexed accumulation or scatter reduction, with means additionally using group counts.
Pair broadcasting uses coordinate maps to gather the projected row and column features before their numerical combination.
Large pair-coordinate maps are constructed on the values device from the per-sample lengths and sample offsets, which are batch-sized, and their construction, gathers and backward accumulation contribute to runtime and temporary storage.
The partitions also provide the segment boundaries consumed by variable-length kernels.

\paragraph{Attention through packed blocks.}
Projections produce token-major queries, keys, and values with static head and feature axes.
The variable-length attention path passes these buffers and cumulative lengths to FlashAttention~\citep{dao2022flashattention,dao2024flashattention2}, and the output adopts the query structure, allowing different query and key lengths in cross-attention.
FlexAttention paths map packed indices to local positions and restrict interactions to matching sample identifiers~\citep{dong2025flexattention}.
Projections, feature normalisations, and residual operations share the packed representation around attention, avoiding conversions at each block boundary.

\paragraph{Convolution over variable-size images.}
Convolution requires neighbourhoods in each image's own coordinate system.
The spatial path gathers sample-local input neighbourhoods, including the receptive-field halo, into batches of equal-size tiles; native convolution processes those tiles and valid outputs are scattered back into packed storage, with coordinates and boundary padding defined within each image so that no interaction crosses a sample boundary.
Backward accumulates overlapping input-gradient contributions and sums parameter gradients across tiles.
This replaces a batch-wide padded envelope with tile buffers, trading padded computation for halo duplication and gather/scatter work; Appendix~\ref{app:implementation} gives the halo geometry and the tiling tradeoff.

\paragraph{Structural costs.}
Execution cost divides into numerical kernels, structural indexing and data movement; packed execution pays off when the saved padded work outweighs the indexing and movement it adds.
Retained partitions additionally let an operator issue one segmented call over the batch instead of one call per sample.

\subsection{Autograd and Dynamic Compilation}
\label{sec:training-integration}

\paragraph{Gradients across representations.}
Gradient connectivity must be preserved where a structured wrapper meets its packed payload.
Autograd records its graph over the payload while the model's forward returns the wrapper, and a compiled block can hand structured singles and pairs to an eager consumer, so that boundary is crossed in both directions within one step and has to be differentiable itself.
A wrapper-to-packed bridge passes the numerical payload to such a consumer and, in backward, rebuilds a gradient carrying the same partitions and dimension order as the forward value.
The inverse bridge attaches packed numerical results to their output wrappers and aligns incoming gradients with the packed dimension order.
Together they preserve the dependency through changes in rank, layout and representation; for Equation~\ref{eq:running-block} they connect the eager loss back to the compiled projections, pair update and reduction.

\paragraph{Runtime sizes and structural schemas.}
Compilation fixes the structural schema at each program point: rank, declared ragged axes, logical dimension order, and static feature layout.
The primary length-dynamic evaluation also fixes batch size; we report no compiled results for traces whose batch size changes.
Different program points may have different schemas: in Equation~\ref{eq:running-block} broadcasting creates a pair and reduction returns singles, so its three program points do not share one schema, though each is fixed across calls.
Per-sample sizes and partitions remain runtime tensor inputs, so supported length-dynamic programs reuse a compiled graph rather than specialising on each length vector.
Output reconstruction combines the fixed schema with these runtime tensors, and custom-operation boundaries represent data-dependent packed extents symbolically.
Explicit declarations preserve an axis's meaning even when observed sizes happen to coincide, so a batch whose lengths are momentarily equal still compiles to the ragged schema.

\section{Experiments}
\label{sec:experiments}
\vspace{-1em}

The evaluation asks when the abstraction improves speed and memory over padding, whether it carries a complete multi-ragged computation, and where the benefit comes from.

\subsection{Setup}
\label{sec:setup}

\paragraph{Workloads and methods.}
The models are randomly initialised systems workloads, measured for execution cost rather than task quality.
The dataset study spans BERT~\citep{devlin2018bert} on IMDB~\citep{maas2011imdb}, GPT-2~\citep{radford2018gpt2} on WikiText-103~\citep{merity2016wikitext}, an encoder--decoder Transformer~\citep{vaswani2017transformer} on WMT14~\citep{bojar2014wmt}, a variable-resolution ViT~\citep{dosovitskiy2021vit} on ImageNet-1K~\citep{russakovsky2015imagenet}, DETR-R50~\citep{carion2020detr} on COCO~\citep{lin2014coco} and FCN-ResNet~\citep{he2016resnet, long2015fcn} on ADE20K~\citep{zhou2019ade20k}; Appendix~\ref{app:protocol} specifies the model scales and configurations.
The main comparison is padded execution, DanLing, and native \code{torch.nested} using its \code{torch.jagged} layout.
Three analysis controls isolate separate effects: comparing attention-only with whole-program packing assesses the benefit of keeping computation packed beyond attention; explicit packing is a direct-programming reference that manages offsets in model code; and matched-granularity execution separates the cost of the tensor interface from the benefit of batch-level call granularity.
Each paired comparison uses the same examples in the same order, with the same initial weights, objective and logical batch partitions.
Eager and Inductor-compiled execution are parallel experiments, each normalised to padding in the same mode.

\paragraph{Measurement.}
The experiments use an A100-SXM4-80GB with PyTorch \code{2.13.0+cu132} and BF16 parameters and floating-point inputs.
TF32 and stochastic layers are disabled for the comparisons.
Host-wall timing covers forward computation, loss, backward and runtime structural work; preparation, input transfer and optimiser updates are outside this compute boundary.
Comparisons against padding use memory efficiency $E_m=M_{\mathrm{pad}}/M_m$ and speedup $S_m=T_{\mathrm{pad}}/T_m$, both improving above one.
Results use one of two windows throughout: \emph{full-trace} pools every measured step, keeping any in-trace compilation in the total, while \emph{steady-state} pools the steps on which \emph{neither} method compiled, so both sides average identical batches.
Time and peak allocated memory are taken over the selected steps; the windows coincide in eager execution.
Figure~\ref{fig:headline} reports steady-state; Figure~\ref{fig:execution} and the result tables report full-trace; Appendix Table~\ref{tab:compilation} gives both where they diverge, and Appendix~\ref{app:implementation} the retained fraction per cell.
Configuration ranges describe variation across model scales rather than independent-run uncertainty.

\paragraph{Validation protocol.}
Values, logical structure and gradients are checked against matching references at the same precision.
The performance matrix retains all completed paired measurements; other outcomes are reported separately using the codes defined in Appendix~\ref{app:protocol}.

\subsection{Speed and Memory Across Workloads}
\label{sec:workloads}

Across the four BERT scales, DanLing gives geometric-mean speedups of \NTIMDBEagerGM{}$\times$ in eager and \NTIMDBCompiledGM{}$\times$ in compiled execution.
BERT-Base uses \NTBertMemorySaving{}\% less peak allocated memory than padding.
The four eager FCN backbones give a \NTADEEagerGM{}$\times$ geometric-mean speedup, with a \NTFCNMemPad{}-to-\NTFCNMemNested{}~GiB allocation reduction for ResNet50.
GPT-2 on WikiText-103 gives an eager geometric-mean speedup of \NTWikiTextEagerGM{}$\times$.
Peak reserved memory favours packing less and, for GPT-2 Small to Large, exceeds padding (Appendix~\ref{app:protocol}).

The vision results separate the two kinds of benefit: variable-resolution ViT has a \NTViTEagerGM{}$\times$ eager geometric-mean throughput ratio while improving allocated-memory efficiency by \NTViTMemGM{}$\times$.
DETR is close to padded throughput (\NTDETRSpeed{}$\times$) with \NTDETRMemEff{}$\times$ memory efficiency, against the matched padded reference that preserves DETR's valid-boundary semantics; Appendix~\ref{app:spatial-baseline}'s conventional padded workflow changes those semantics and runs faster still.

Eager WMT-Base saves substantial memory but runs at \NTWMTEagerRatio{}$\times$ padded throughput.
Its compiled speedup is \NTWMTSteadySpeed{}$\times$ in steady state and \NTWMTTraceSpeed{}$\times$ over the full trace, reflecting \NTWMTRecompiles{} additional padded recompilations that DanLing's fixed compiled schema avoids.
Native jagged reaches steady-state throughput close to DanLing's in the completed compiled GPT-2 and WMT14 cells, while their eager results differ more widely across workloads.

\subsection{Composing Multiple Ragged Axes}
\label{sec:composition-training}

The single--pair--single program of Equation~\ref{eq:running-block} directly exercises the paper's central design, with both of its outputs used by an eager, sample-weighted squared loss.
The same program is expressed in square, rectangular and non-leading layouts, with $B=8$, single width 384 and pair width 128.
Padded, explicit-packed and DanLing executions share the feature transformations and objective.

\begin{table}[t]
  \centering
  \caption{\textbf{Single--pair--single program, eager and compiled.} Full-trace mean ms/step / peak allocated GiB; the final column is speedup over padding. Both modes run the same logical workloads.}
  \label{tab:mechanism}
  \begin{tabular*}{\linewidth}{@{\extracolsep{\fill}}llrrrr@{}}
\toprule
Layout & Mode & Padded & Explicit packed & DanLing & Speedup \\
\midrule
square & eager & 25.1 / 5.59 & 8.7 / 1.60 & 11.1 / 1.61 & 2.26$\times$ \\
square & compiled & 21.2 / 4.64 & 7.6 / 1.35 & 9.5 / 1.25 & 2.23$\times$ \\
rectangular & eager & 24.2 / 4.99 & 9.1 / 1.62 & 11.6 / 1.62 & 2.09$\times$ \\
rectangular & compiled & 18.1 / 4.14 & 8.0 / 1.36 & 10.1 / 1.26 & 1.79$\times$ \\
nonleading & eager & 25.7 / 5.53 & 8.5 / 1.63 & 11.4 / 1.63 & 2.26$\times$ \\
nonleading & compiled & 22.1 / 4.59 & 7.5 / 1.37 & 9.8 / 1.27 & 2.25$\times$ \\
\bottomrule
\end{tabular*}

\end{table}

NestedTensor runs \NTMechanismSpeedMin{}--\NTMechanismSpeedMax{}$\times$ faster than padding across these layouts and modes (Table~\ref{tab:mechanism}).
The compiled program reuses its graph across the tested changing partitions and extents.
NestedTensor and explicit packing both run at whole-batch granularity here, the latter through gather indices and a segmented reduction; on this program of cheap operators NestedTensor's latency is \NTMechanismOverheadMin{}--\NTMechanismOverheadMax{}\% higher, whereas the matched-granularity Pairformer control of Section~\ref{sec:analysis} observes at most \NTPairIfaceCostMax{}\% on fused kernels.

\paragraph{A larger pair application.}
A four-block Pairformer-style workload extends the comparison to repeated triangle multiplication, starting- and ending-node triangle attention, single attention and feature transitions~\citep{abramson2024alphafold3}, at single width 384 and pair width 128.
Padding runs the triangle updates as native \code{einsum} contractions and masked attention over the envelope.
For square pairs, explicit packing, which again manages offsets itself, invokes OOps (a fused triangle-kernel library) dense operators per sample, while NestedTensor invokes the corresponding segmented operators over the batch; both run pair-biased single attention per sample on unpacked slices.

\begin{table}[t]
  \centering
  \caption{\textbf{Four-block Pairformer-style application, eager execution.}
  Full-trace mean ms/step / peak allocated GiB; the final column is speedup over padding. The square regimes span length variability; padding runs native PyTorch kernels.}
  \label{tab:pairformer}
  \adjustbox{max width=\linewidth}{\begin{tabular}{@{}lrrrr@{}}
\toprule
Shape regime & Padded & Explicit packed & DanLing & Speedup \\
\midrule
Square, near-uniform ($B=8$) & 634.2 / 38.09 & 397.9 / 17.81 & 264.5 / 17.43 & 2.40$\times$ \\
Square, moderate ($B=8$) & 588.2 / 35.02 & 403.0 / 12.97 & 212.0 / 12.78 & 2.77$\times$ \\
Square, high variation ($B=8$) & 641.8 / 38.08 & 396.4 / 5.44 & 148.7 / 5.41 & 4.32$\times$ \\
Square, extreme variation ($B=16$) & 638.8 / 34.76 & 802.7 / 3.04 & 203.7 / 3.06 & 3.14$\times$ \\
Non-leading ($B=8$) & 651.5 / 33.12 & 488.5 / 22.83 & 494.6 / 22.83 & 1.32$\times$ \\
Rectangular ($B=8$) & 650.6 / 38.43 & 542.9 / 23.93 & 547.8 / 23.93 & 1.19$\times$ \\
\bottomrule
\end{tabular}
}
\end{table}

The measured square regimes give \NTPairPadMin{}--\NTPairPadMax{}$\times$ speedups over padding, which combine packing, fused kernels and call granularity; explicit-packed and NestedTensor allocations are similar to each other and both well below padded.
The non-leading and rectangular variants run per sample in both packed implementations, which then differ by no more than the interface overhead of Section~\ref{sec:analysis}.

\paragraph{Compiled execution.}
Under the default Inductor configuration, DanLing completes compiled execution in all four square regimes, while the tested padded and explicit-packed implementations fail during compilation.
They fail at different points: padded execution fails an Inductor divisibility check on a symbolic pair extent, and explicit packing fails fake-tensor propagation on its data-dependent gather indices (Appendix~\ref{app:implementation}).
Explicit packing's failure is the kind the runtime-tensor metadata of Section~\ref{sec:training-integration} avoids; padding's is a limitation of Inductor's fusion.
The complete pair application therefore compiles, not only the single--pair--single block of Equation~\ref{eq:running-block}.
Disabling Inductor fusion entirely, though not epilogue fusion alone, lets the padded implementation compile as a full graph, but only \NTPairFusionOffEagerMin{}--\NTPairFusionOffEagerMax{}$\times$ faster than its eager execution; DanLing compiled is \NTPairFusionOffDanLingMin{}--\NTPairFusionOffDanLingMax{}$\times$ faster again, a gap combining padding removal with fusion and kernel choice.
Relative to its own eager execution, DanLing is \NTPairCompiledMin{}--\NTPairCompiledMax{}$\times$ faster in steady state, with \NTPairCompiledRecompiles{} of the four regimes recompiling once during the measured trace.

\subsection{Performance Analysis}
\label{sec:analysis}

\begin{figure}[t]
  \centering
  \includegraphics[width=\linewidth]{\NTArtifactRoot/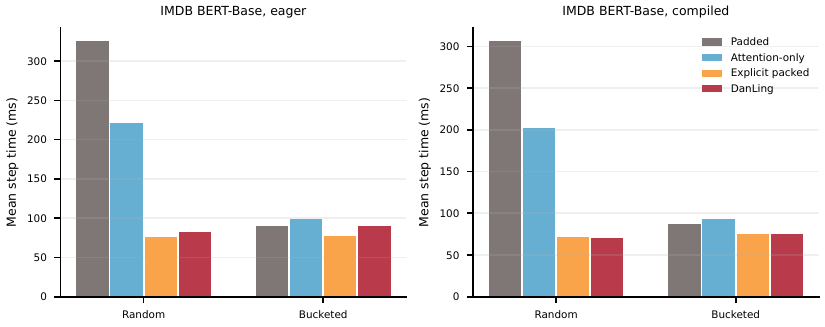}
  \caption{\textbf{Packing attention versus packing the complete program.}
    The four controls use the same IMDB BERT-Base workload.
    Attention-only packing includes unpadding and repadding; explicit packing and DanLing keep surrounding operations packed.
  Eager and compiled panels use full-trace mean step time (Section~\ref{sec:setup}).}
  \label{fig:execution}
\end{figure}

\paragraph{Whole-program packing.}
Figure~\ref{fig:execution} separates four implementations of the same BERT workload, from padding the whole model to keeping every surrounding operation packed.

For random batching, replacing padded attention helps, but the larger gain comes from also keeping the surrounding feature computation packed.
Attention-only packing uses DanLing's variable-length backend with every other operation left padded, so it separates the backend from the rest: \NTCtlAttnEager{}$\times$ padded throughput eager and \NTCtlAttnCompiled{}$\times$ compiled, against DanLing's \NTCtlDanLingEager{}$\times$ and \NTCtlDanLingCompiled{}$\times$.
In compiled execution, DanLing and explicit packing differ by at most \NTCtlCompiledGap{}\% in full-trace throughput under either batching policy, so a tensor interface can approach direct packed performance without model-level offset bookkeeping; eager execution retains a larger gap, as the multi-ragged program also shows.

\paragraph{Separating interface cost from execution strategy.}
Once padded cells are gone, how the remaining work is issued still matters.
On the Pairformer workload NestedTensor is \NTPairPackedMin{}--\NTPairPackedMax{}$\times$ faster than explicit packing, the opposite direction to the single--pair--single program of Section~\ref{sec:composition-training}, because that baseline also differs in how it calls the kernels.
A third implementation separates the two effects, keeping the NestedTensor interface but issuing one dense call per sample as the baseline does.
At matched granularity the interface adds at most \NTPairIfaceCostMax{}\% to latency across the four square regimes, while moving to one segmented call per operator over the whole batch gives \NTPairGranMin{}--\NTPairGranMax{}$\times$ (Appendix Table~\ref{tab:attribution}); the speedups above are that gain net of the interface overhead, up to run-to-run variation.

\paragraph{Batching policy.}
How much redundant work exists to remove depends on the batch: for sequences the valid fraction is $\rho_{\mathrm{seq}}=\sum_iN_i/BN_{\max}$, with the image and pair analogues in Appendix~\ref{app:protocol}.
Random and bucketed IMDB traces hold the same examples, partitioned differently, and bucketing raises token occupancy from \NTIMDBOccRandom{}\% to \NTIMDBOccBucket{}\%.
The eager DanLing/padded throughput ratio then changes from \NTCtlDanLingEager{}$\times$ to \NTCtlBucketEager{}$\times$, and the compiled ratio from \NTCtlDanLingCompiled{}$\times$ to \NTCtlBucketCompiled{}$\times$.
Packing and bucketing are therefore complementary: the representation preserves the chosen batch, while the gain available depends on that batch's geometry.

\paragraph{Model scale and shape variability.}
BERT speedups increase with model scale before levelling off at Base and Large, while length heterogeneity moves a different workload: across the square Pairformer regimes at $B=8$ the speedup over padding rises from \NTPairPadMin{}$\times$ under near-uniform lengths to \NTPairPadMax{}$\times$ under high variability, because padding's cost tracks the batch maximum whatever the distribution.
At that near-uniform end the lengths span only \NTPairUniformLengths{} and pair occupancy is \NTPairUniformOcc{}\% (Appendix Table~\ref{tab:regimes}), leaving little pair-cell padding to remove; its speedup combines packing, kernel choice and call granularity, which this comparison does not separate.
The extreme regime additionally doubles the batch to $B=16$ and is reported separately (Table~\ref{tab:pairformer}; Appendix Figure~\ref{fig:scaling}).

\section{Discussion and Conclusion}
\label{sec:conclusion}
\vspace{-1em}

NestedTensor keeps logical structure available after packing and transforms that structure together with the numerical values, through a common PyTorch tensor interface.
The complete four-block pair application compiles and trains under this representation with default compiler settings where the padded and hand-packed implementations we measured do not, although the simpler single--pair--single block compiles for all three.
The measurements then separate where the efficiency comes from: on BERT, packing the whole program rather than attention alone; on the pair workload, issuing batch-level segmented calls rather than per-sample ones.
How much there is to gain stays conditional---bucketing recovers most of it on IMDB, and memory savings can coexist with slower execution---but the composition it rests on does not.

Structure that survives into execution is useful twice: it removes padded work, and it leaves the execution layer enough information to choose how the remaining work is issued.

\label{page:main-end}
\clearpage

\subsection*{AI use statement}
Generative AI assistants were used to support software development, experiment tooling, analysis, and manuscript preparation.
They were not used for research ideation or for literature search and related-work discovery: the research question, the design of the representation and its operator rules, and the choice and framing of the experiments are the author's own, and every cited work was located and read by the author.
No reported quantity was produced by a language model: every number in the text, tables and figures is emitted by one script from the frozen benchmark records.
AI-assisted code was reviewed by running it and comparing its output against a reference implementation; AI-assisted text was checked against the stored artifacts, and its cross-references, macros and citations verified, by automated scans over the sources.
The author reviewed all AI-assisted work and takes responsibility for the implementation, analysis, claims, and final manuscript.

\subsection*{Reproducibility statement}
The representation and its structural transformations are specified in Section~\ref{sec:method}, with reference semantics in Appendix~\ref{app:semantics} and execution paths in Appendix~\ref{app:implementation}.
Appendix~\ref{app:protocol} records the environment, model configurations, measurement boundaries and the outcome codes used when a cell has no completed measurement; Section~\ref{sec:setup} defines the two timing windows, and Appendix~\ref{app:implementation} reports how many steps the steady-state window retains.

\subsection*{Ethics statement}
The experiments use existing public NLP and vision datasets under their respective access terms and licenses.
They measure execution efficiency without collecting new human-subject data or deploying trained models.
The models are randomly initialised, so no model capable of downstream use is produced or released.

\bibliography{bibliography}
\bibliographystyle{iclr2027_conference}

\appendix

\section{Representation, Semantics, and the Running Program}
\label{app:semantics}
\vspace{-1em}

\paragraph{Structural invariants.}
Ragged axes in this work vary between individually dense elements of a common rank.
The representation retains static feature axes, an ordered tuple of ragged axes, and the batch-position convention.
For the batch-leading exposition, element axis $d$ is public axis $d+1$.
The implementation also supports its sequence-first convention; the batch axis is not an arbitrary permutable element dimension.
Sample offsets satisfy $o_0=0$, $o_{i+1}-o_i=p_i$, and $o_B=T$.
Zero-volume elements may have coincident consecutive offsets, while the exact size matrix distinguishes their remaining logical dimensions.

Sizes and offsets play different roles.
For example, element shapes $(2,6,C)$ and $(3,4,C)$ both occupy twelve packed feature rows but cannot be aligned as the same logical pair tensor.
Hierarchical row splits identify the groups consumed by reductions.
For rectangular elements $(N_i,M_i,C)$, the innermost splits contain $1+\sum_iN_i$ entries; they need not be $O(B)$ metadata.
Canonical sizes and offsets stay on the CPU, with device mirrors and execution indices used by GPU consumers.

\paragraph{Reference semantics.}
Let $\mathcal D$ denote the logical interpretation of a structured tensor; it acts componentwise on argument tuples and leaves dense tensors unchanged.
A deterministic functional handler $\widehat f$ is specified against its reference $f_{\mathrm{ref}}$ by $\mathcal D(\widehat f(\mathbf Z))=f_{\mathrm{ref}}(\mathcal D(\mathbf Z))$, the contract cited in Section~\ref{sec:dispatch}.
Sample-local operations are compared with independent dense element computations using the same weights, masks, coordinate conventions, and loss weighting.
Global reductions and operations combining samples require their corresponding batch reference.
A mean over all stored entries weights entries equally.
The supplied API's single batch-axis reduction instead performs a complete reduction within each element and stacks the results: for elements $[1,2]$ and $[3,4]$, its \code{sum(dim=0)} gives $[3,7]$.
This convention is distinct from a dense column-wise batch reduction.
The cross-entropy handler uses a trailing class dimension, so a reference must align the objective accordingly.

The validity mask uses true for valid positions, while attention interfaces can use different boolean conventions.
The adapter translates those conventions and combines padding validity with model masks.
Padded accessors are interoperability operations; the canonical numerical payload remains packed.
That contract is a specification; derivative and structural agreement are tested and reported separately.

\subsection{Measured single--pair--single program}
\label{app:running-program}

The measured mechanism uses the projections, residual pre-normalised SwiGLU update, and column mean in Equation~\ref{eq:running-block}.
Pair construction is the sum of the two projected single features; there is no additional activation at that construction step.
Both returned outputs are consumed outside the compiled forward by
\begin{equation}
 \mathcal L =
 \frac{\sum_i\alpha_i\|\widetilde X_i\|_F^2}{S C\sum_i N_i}
 +\frac{\sum_i\beta_i\|\widetilde P_i\|_F^2}{S C_p\sum_i N_iM_i},
 \qquad \alpha_i,\beta_i>0.
 \label{eq:joint-diagnostic-loss}
\end{equation}
Here $M_i=N_i$ for square pairs, $S=1$ for leading layouts, and $S=3$ for the non-leading variant.
The rectangular variant has distinct row and column inputs and divides its column mean by $M_i$.
The sample weights are $\alpha_i=1+i/B$ and $\beta_i=1.5-i/(2B)$ for $i=0,\ldots,B-1$.
The sample weights and losses are identical across representations; the squared sums accumulate in at least FP32.
This mechanism has no trainable readout after the returned pair state.
The separate Pairformer application does include a trainable pair readout in its eager loss.

The mechanism uses $B=8$, $C=384$, $C_p=128$, and transition expansion four.
Eight validation patterns cover changed partitions with fixed totals, changed packed extents and maxima, and fresh equal structures.
Each measured method/layout/mode cell has three traces of two steps.
All eight patterns pass the recorded structure, output, gradient and one-update checks for all 18 combinations.

\section{Implementation and Compilation Scope}
\label{app:implementation}
\vspace{-1em}

\paragraph{Execution paths.}
Operator availability, a packed implementation, and fullgraph compilation are distinct properties.
The two registries hold \NTOperatorEntries{} keys for \NTOperatorCount{} distinct operators, since one operator owns several keys, as \path{torch.add}, \path{Tensor.add}, \path{Tensor.__add__} and \path{aten::add.Tensor} do; operators are counted after dropping module paths and overload suffixes, and an in-place operator counts separately from its out-of-place form because it mutates packed storage through its own handler.
Feature-wise operations reuse dense packed kernels.
Ragged reductions and broadcasts use coordinate/group maps; selected attention paths use native variable-length or FlexAttention execution.
Spatial paths gather tile neighbourhoods and invoke dense convolution, or use direct pooling kernels; selected pooling paths instead address packed pixels directly through per-sample shapes and offsets.
Pointwise convolution reuses the dense packed path after aligning the channel axis with the static feature dimension.
For output tile dimensions $(t_h,t_w)$, stride $s$, dilation $\delta$ and kernel size $k$, the gathered input tile has dimensions $u_h=(t_h-1)s_h+\delta_h(k_h-1)+1$ and $u_w=(t_w-1)s_w+\delta_w(k_w-1)+1$.
Tile size therefore balances numerical batching against overlapping halo reads, boundary tiles and gather/scatter work; chunking bounds the number of simultaneous tile buffers, the dense backend may use channels-last storage, and backward can regenerate input tiles from the saved packed input for weight gradients.
Some bindings, including segmented \code{cdist} and \code{cumprod}, retain per-sample native kernel invocations behind a custom-operation boundary.
Eager fallback and unsupported compiled argument combinations remain part of the implementation's declared scope; a workload counts as passing only on its own measured outcome, never on operator availability.

\paragraph{Dynamic metadata and differentiation.}
The principal length-dynamic contract fixes batch size and the structural schema at each program point.
The flattening protocol exposes one differentiable payload child alongside nondifferentiable structural metadata.
Tensor-backed sizes and row splits are runtime inputs; standalone FakeTensor construction retains the Python metadata needed where symbolic data-dependent extents are unavailable.
Eager reconstructions propagate weak dynamic annotations before entering a compiled region.
This prevents a ragged extent from being accidentally identified with an unrelated metadata dimension of the same observed size.
The optimised eager propagation recognises the state installed by the public marker and preserves a public-API path for missing marker metadata, tensor subclasses, and FakeTensor execution.

Wrapper-to-packed and packed-to-wrapper autograd bridges preserve the corresponding numerical dependency and packed dimension order, so changing partitions reuse the compiled graph for the evaluated program and gradients remain connected across the compiled/eager boundary.

\paragraph{Current compiled application outcomes.}
The ordinary-operation mechanism completes in all three layouts.
The dataset matrix records upstream native-jagged BERT guard errors, failed padded vision compilation, and NestedTensor FCN fake-propagation failures.
Under the default Inductor configuration the full Pairformer compiles for DanLing but not for either baseline (Section~\ref{sec:composition-training}).
The recorded failures are identical across the four square regimes: padded raises \code{InductorError} with \code{CantSplit: 128*s83 not divisible by ((s83**2)//s83)}, and explicit packing raises \code{TorchRuntimeError} during fake-tensor propagation.
The padded failure has no user frame: Inductor's scheduler raises it while fusing nodes whose iteration extents are $s^2$ and $128s$, a constraint that holds for every positive integer $s$ but that it cannot establish.
Padded therefore compiles under static shapes, and under dynamic shapes once fusion is disabled (\code{max_fusion_size=1}; \code{epilogue_fusion=False} alone still fails), running at \NTPairFusionOffMsMin{}--\NTPairFusionOffMsMax{}~ms/step in one harness whose padded eager reproduces Table~\ref{tab:pairformer}.
Explicit packing fails before any fusion decision and was not retested.
Of DanLing's four compiled regimes, \NTPairCompiledRecompiles{} recompile once inside the measured trace, unlike the sequence workloads, where the compiled schema is reused throughout.
For FCN, padded ResNet18/101 reach the recompilation limit, ResNet152 exhausts the recorded two-hour budget without a completed outcome, and the standalone ResNet50 padded compiled cell has no result artifact.
The four NestedTensor FCN compiled cells fail during convolution fake propagation.

\paragraph{Steady-state execution versus in-trace recompilation.}
A measured step whose shape was not already covered by a compiled graph triggers a new frontend/backend capture; eager execution has no such event, so the distinction only matters for compiled runs.
Figure~\ref{fig:headline}'s compiled-mode speedup pools only steps that did not themselves trigger a capture, isolating kernel and packing efficiency from compiler behaviour.
Every steady-state ratio pools the steps on which neither method compiled, so numerator and denominator average identical batches.
That intersection retains all paired steps except in \NTRetainedCells{} compiled cells: \NTKeptWikiText{} of \NTTotalWikiText{} for WikiText Small, Medium and Large, \NTKeptWMTBase{} of \NTTotalWMTBase{} for WMT14-Base, \NTKeptWMTBig{} of \NTTotalWMTBig{} for WMT14-Big, and \NTKeptWikiTextXL{} of \NTTotalWikiTextXL{} for WikiText-XL, whose short trace makes it the only cell where the restriction is material (it moves the ratio by \NTRetainShiftXL{}\%; elsewhere the shift is at most \NTRetainShiftOther{}\%).
Table~\ref{tab:compilation} reports the complement for the compiled cells whose full-trace and steady-state ratios diverge most (Section~\ref{sec:workloads}): how many graphs each method compiled over its lifetime, including calibration, and how many of those compilations were shape-driven recompilations occurring after calibration, alongside the resulting steady-state and full-trace per-step times.
In these recorded traces the padded implementations recompile more often than DanLing and native jagged, which reuse one compiled schema in \NTCompileReuseCells{} of the \NTCompileCells{} cells; WikiText-XL is the exception, where its \NTXLSteps{}-step trace leaves every method recompiling \NTXLRecompileMin{} or \NTXLRecompileMax{}.

\begin{table}[htbp]
\centering
\caption{\textbf{Dynamic-compilation evidence.} ``Graphs'' counts lifetime frontend/backend invocations, including calibration; ``recompile'' counts only those occurring after calibration, during the measured trace. ``Compile step'' is the mean measured time of a recompiling step, execution and compilation combined; it is not an isolated compiler stopwatch. Steady-state time pools non-recompiling steps; full-trace time pools every measured step.}
\label{tab:compilation}
\adjustbox{max width=\linewidth}{\begin{tabular}{@{}llrrrrr@{}}
\toprule
Workload & Method & Graphs & Recompiles & Compile step (ms) & Steady (ms) & Full trace (ms) \\
\midrule
WMT14 / Base & Padded & 8 & 4 & 49567 & 43.7 & 1591.3 \\
 & Native jagged & 2 & 0 & -- & 28.9 & 28.9 \\
 & DanLing & 2 & 0 & -- & 27.4 & 27.4 \\
WMT14 / Big & Padded & 6 & 3 & 56811 & 87.8 & 1417.3 \\
 & Native jagged & 2 & 0 & -- & 38.7 & 38.7 \\
 & DanLing & 2 & 0 & -- & 39.1 & 39.1 \\
WikiText / Small & Padded & 3 & 1 & 28563 & 200.7 & 643.9 \\
 & Native jagged & 2 & 0 & -- & 138.4 & 138.4 \\
 & DanLing & 2 & 0 & -- & 138.3 & 138.3 \\
WikiText / Medium & Padded & 3 & 1 & 54697 & 504.1 & 1350.9 \\
 & Native jagged & 2 & 0 & -- & 353.5 & 353.5 \\
 & DanLing & 2 & 0 & -- & 362.1 & 362.1 \\
WikiText / Large & Padded & 4 & 1 & 106792 & 997.4 & 2650.4 \\
 & Native jagged & 2 & 0 & -- & 701.8 & 701.8 \\
 & DanLing & 2 & 0 & -- & 709.9 & 709.9 \\
WikiText / XL & Padded & 5 & 2 & 128865 & 454.2 & 43257.6 \\
 & Native jagged & 4 & 1 & 96345 & 444.8 & 16428.1 \\
 & DanLing & 4 & 1 & 153093 & 444.6 & 25886.1 \\
\bottomrule
\end{tabular}
}
\end{table}

\section{Benchmark Protocol and Complete Dataset Matrix}
\label{app:protocol}
\vspace{-1em}

\paragraph{Environment and source cohorts.}
The PyTorch build is \code{2.13.0+cu132}, commit
\code{cf30153c4c131c8164ee7798e5022d810682e2cb}, with CUDA~13.2 and Triton~3.7.1 on A100-SXM4-80GB GPUs.
Parameters use BF16; TF32 is disabled, float32 matmul precision is \code{highest}, and stochastic layers are disabled for these comparisons.
The validation optimiser is SGD at $10^{-5}$, with BF16 parameters and no separate FP32 master-weight copy.

The reported tables and figures are not all measurements of a single checkout: the dataset and execution-control timings, the mechanism snapshot, and the square Pairformer timings were each frozen at a different source revision.
Every exported table and figure is tied to frozen manifests recording the exact source revision and input-file hashes, distributed with the artifact.
The Pairformer cells additionally depend on the OOps kernel library, whose public release postdates these measurements.

\begin{figure}[htbp]
\centering
\includegraphics[width=\linewidth]{\NTArtifactRoot/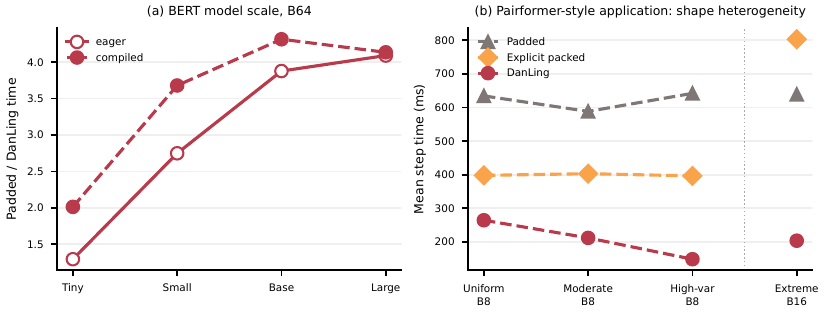}
\caption{\textbf{Model-scale and shape-heterogeneity sensitivity.}
\textbf{(a)} BERT model-scale speedup (padded / DanLing mean step time) at $B=64$, eager (hollow) and compiled (solid).
\textbf{(b)} Eager mean step time for the same four-block Pairformer-style workload as Table~\ref{tab:pairformer}, for padded, explicit-packed and DanLing across four length-heterogeneity regimes at fixed model size, from nearly uniform lengths to one long chain among many short ones; the rightmost regime additionally doubles the batch to $B=16$.
Every point compares methods within the same execution mode.}
\label{fig:scaling}
\end{figure}

The image and pair occupancy fractions accompanying $\rho_{\mathrm{seq}}$ in Section~\ref{sec:analysis} are $\rho_{\mathrm{image}}=\sum_iH_iW_i/BH_{\max}W_{\max}$ and $\rho_{\mathrm{pair}}=\sum_iN_iM_i/BN_{\max}M_{\max}$.
Table~\ref{tab:regimes} lists the lengths and $\rho_{\mathrm{pair}}$ behind Tables~\ref{tab:mechanism} and~\ref{tab:pairformer}.

\begin{table}[htbp]
\centering
\caption{\textbf{Length regimes of the pair workloads.} Lengths span every measured sample (rows $\times$ columns for rectangular pairs); $\rho_{\mathrm{pair}}$ pools valid and padded pair cells over the measured batches.
A cell's eager and compiled runs, and the matched-granularity run of a square regime (Table~\ref{tab:attribution}), use the same length schedule.}
\label{tab:regimes}
\adjustbox{max width=\linewidth}{\begin{tabular}{@{}llrr@{}}
\toprule
Workload & Shape regime & Lengths & $\rho_{\mathrm{pair}}$ \\
\midrule
Single--pair--single & Square ($B=8$) & 66--317 & 27.2\% \\
Single--pair--single & Rectangular ($B=8$) & 66--307 $\times$ 66--309 & 29.6\% \\
Single--pair--single & Non-leading ($B=8$) & 40--182 & 28.0\% \\
Pairformer-style & Square, near-uniform ($B=8$) & 208--224 & 94.4\% \\
Pairformer-style & Square, moderate ($B=8$) & 137--216 & 71.6\% \\
Pairformer-style & Square, high variation ($B=8$) & 88--224 & 27.9\% \\
Pairformer-style & Square, extreme variation ($B=16$) & 44--160 & 14.8\% \\
Pairformer-style & Non-leading ($B=8$) & 90--130 & 68.5\% \\
Pairformer-style & Rectangular ($B=8$) & 130--221 $\times$ 148--224 & 66.2\% \\
\bottomrule
\end{tabular}
}
\end{table}

\paragraph{Architecture and batch sizes.}
IMDB uses BERT Tiny/Small configurations from \citet{turc2019well} and Base/Large configurations from \citet{devlin2018bert}, with $(L,H,C)$ equal to $(2,2,128)$, $(4,8,512)$, $(12,12,768)$, and $(24,16,1024)$, all at $B=64$.
WikiText uses GPT-2 Small/Medium/Large/XL shapes with $(L,H,C)$ equal to $(12,12,768)$, $(24,16,1024)$, $(36,20,1280)$, and $(48,25,1600)$.
Its main document-length cap is 4096, with $B=8$ except XL at $B=2$.
The WMT Base/Big settings use six encoder and six decoder layers, widths 512/1024, and 8/16 heads at $B=64$.
ViT Ti/S use the DeiT-Ti/S scale configurations of \citet{touvron2021deit}, while B/L follow \citet{dosovitskiy2021vit}; all use patch size 16, widths 192/384/768/1024, depths 12/12/12/24, and 3/6/12/16 heads at $B=32$.
DETR uses ResNet50, six encoder and six decoder layers, hidden width 256, eight heads, and 100 queries at $B=32$.
FCN uses ResNet18/50/101/152 and 150 classes at $B=32$.

\paragraph{Measurement accounting.}
Timing follows PyTorch's benchmarking and CUDA synchronisation guidance~\citep{pytorchBenchmark,pytorchCUDASemantics}.
Every cell fixes one execution mode.
Batch identities and sizes come from the frozen sibling manifest, not placeholder configuration lengths or a mutable external manifest.
Completed traces are checked for matching IDs, repetition IDs and step counts.
Latency sums all measured steps; throughput divides all measured examples by that elapsed time.
Memory takes the maximum allocated or reserved bytes across all measured traces.
No trace is removed for containing compilation.
Validation and calibration timings are outside the compute measurement, and their cache priming means that a timed capture is not a measurement of all compilation costs from a pristine process.
The primary tables record one repetition and descriptive step variation.
Inference about independent-run uncertainty or maximum feasible batch size is outside these records.

\paragraph{Execution outcomes.}
Where a cell has no completed measurement, \textsc{nd} marks no run on record, \textsc{bl} marks a padded reference that itself failed validation (blocking the comparison), \textsc{g} marks a dynamic-shape guard error during compiled dispatch, \textsc{cf} marks any other compilation failure, \textsc{nr} marks a run that ended without a completed outcome, and \textsc{ne} marks a method not evaluated because the workload falls outside its structural support (e.g.\ native jagged on COCO/ADE20K); unfinished measurements keep \textsc{nr} rather than a diagnosed cause.
The complete grids below report each configuration as two rows: mean throughput in samples/s, then peak allocated / reserved memory in GiB.
Reserved memory additionally counts blocks the caching allocator holds for reuse, and favours packing less than allocated memory: for GPT-2 Small to Large, DanLing and native jagged reserve more than padding.

\subsection{IMDB / BERT}
\begin{center}\adjustbox{max width=\linewidth}{%
\begin{tabular}{@{}ll l rrr@{}}\toprule
Size & Mode & & Padded & Native jagged & DanLing \\ \midrule
tiny & eager & samples/s & 4635.7 & 1651.7 & 5990.2 \\
 &  & GiB alloc/res. & 0.67 / 0.82 & 0.29 / 0.63 & 0.29 / 0.63 \\
tiny & compiled & samples/s & 6625.6 & \textsc{g} & 13330.0 \\
 &  & GiB alloc/res. & 0.61 / 0.64 & -- & 0.24 / 0.43 \\
small & eager & samples/s & 932.3 & 996.2 & 2562.6 \\
 &  & GiB alloc/res. & 4.50 / 4.66 & 1.76 / 4.05 & 1.76 / 4.05 \\
small & compiled & samples/s & 1036.1 & \textsc{g} & 3813.8 \\
 &  & GiB alloc/res. & 4.40 / 4.48 & -- & 1.67 / 3.91 \\
base & eager & samples/s & 197.1 & 442.0 & 764.8 \\
 &  & GiB alloc/res. & 18.85 / 19.06 & 7.17 / 16.23 & 7.17 / 16.23 \\
base & compiled & samples/s & 208.9 & \textsc{g} & 901.7 \\
 &  & GiB alloc/res. & 18.75 / 18.86 & -- & 7.07 / 16.12 \\
large & eager & samples/s & 67.1 & 231.0 & 274.7 \\
 &  & GiB alloc/res. & 49.50 / 49.65 & 18.72 / 42.45 & 18.72 / 42.55 \\
large & compiled & samples/s & 69.8 & \textsc{g} & 289.0 \\
 &  & GiB alloc/res. & 49.39 / 49.54 & -- & 18.61 / 42.34 \\
\bottomrule\end{tabular}}\end{center}

\subsection{WikiText-103 / GPT-2}
\begin{center}\adjustbox{max width=\linewidth}{%
\begin{tabular}{@{}ll l rrr@{}}\toprule
Size & Mode & & Padded & Native jagged & DanLing \\ \midrule
small & eager & samples/s & 37.4 & 46.4 & 54.4 \\
 &  & GiB alloc/res. & 22.44 / 22.45 & 19.08 / 39.21 & 19.08 / 39.21 \\
small & compiled & samples/s & 12.4 & 57.8 & 57.8 \\
 &  & GiB alloc/res. & 22.43 / 26.21 & 19.07 / 39.20 & 19.07 / 39.20 \\
medium & eager & samples/s & 15.2 & 21.0 & 21.9 \\
 &  & GiB alloc/res. & 37.93 / 37.95 & 32.29 / 53.00 & 32.29 / 53.00 \\
medium & compiled & samples/s & 5.9 & 22.6 & 22.1 \\
 &  & GiB alloc/res. & 37.93 / 41.73 & 32.29 / 52.37 & 32.29 / 52.37 \\
large & eager & samples/s & 7.7 & 11.0 & 11.2 \\
 &  & GiB alloc/res. & 59.80 / 63.54 & 50.97 / 75.59 & 50.97 / 75.59 \\
large & compiled & samples/s & 3.0 & 11.4 & 11.3 \\
 &  & GiB alloc/res. & 59.80 / 59.88 & 50.97 / 76.00 & 50.97 / 76.00 \\
xl & eager & samples/s & 4.6 & 3.5 & 4.6 \\
 &  & GiB alloc/res. & 25.31 / 35.16 & 25.31 / 30.87 & 25.31 / 30.87 \\
xl & compiled & samples/s & 0.046 & 0.12 & 0.077 \\
 &  & GiB alloc/res. & 25.31 / 35.25 & 25.31 / 30.95 & 25.31 / 30.95 \\
\bottomrule\end{tabular}}\end{center}

\subsection{WMT14 / Transformer}
\begin{center}\adjustbox{max width=\linewidth}{%
\begin{tabular}{@{}ll l rrr@{}}\toprule
Size & Mode & & Padded & Native jagged & DanLing \\ \midrule
base & eager & samples/s & 1322.0 & 335.4 & 884.2 \\
 &  & GiB alloc/res. & 12.71 / 39.89 & 2.81 / 4.88 & 2.81 / 4.88 \\
base & compiled & samples/s & 40.2 & 2214.8 & 2339.9 \\
 &  & GiB alloc/res. & 12.70 / 39.96 & 2.81 / 7.29 & 2.81 / 7.29 \\
big & eager & samples/s & 702.7 & 335.6 & 875.1 \\
 &  & GiB alloc/res. & 16.92 / 42.41 & 3.94 / 8.39 & 3.94 / 8.39 \\
big & compiled & samples/s & 45.2 & 1652.2 & 1636.4 \\
 &  & GiB alloc/res. & 16.89 / 42.44 & 3.95 / 8.37 & 3.95 / 8.32 \\
\bottomrule\end{tabular}}\end{center}

\subsection{ImageNet / ViT}
\begin{center}\adjustbox{max width=\linewidth}{%
\begin{tabular}{@{}ll l rrr@{}}\toprule
Size & Mode & & Padded & Native jagged & DanLing \\ \midrule
ti & eager & samples/s & 498.8 & 202.1 & 351.3 \\
 &  & GiB alloc/res. & 1.46 / 2.01 & 0.97 / 1.21 & 0.97 / 1.23 \\
ti & compiled & samples/s & \textsc{bl} & \textsc{bl} & \textsc{bl} \\
 &  & GiB alloc/res. & -- & -- & -- \\
s & eager & samples/s & 490.1 & 202.7 & 343.4 \\
 &  & GiB alloc/res. & 2.82 / 3.02 & 1.90 / 3.11 & 1.90 / 3.11 \\
s & compiled & samples/s & \textsc{bl} & \textsc{bl} & \textsc{bl} \\
 &  & GiB alloc/res. & -- & -- & -- \\
base & eager & samples/s & 341.6 & 199.6 & 343.7 \\
 &  & GiB alloc/res. & 5.65 / 6.06 & 3.84 / 6.17 & 3.84 / 6.17 \\
base & compiled & samples/s & \textsc{bl} & \textsc{bl} & \textsc{bl} \\
 &  & GiB alloc/res. & -- & -- & -- \\
l & eager & samples/s & 135.4 & 110.5 & 184.2 \\
 &  & GiB alloc/res. & 14.49 / 15.14 & 10.04 / 15.88 & 10.04 / 15.88 \\
l & compiled & samples/s & \textsc{bl} & \textsc{bl} & \textsc{bl} \\
 &  & GiB alloc/res. & -- & -- & -- \\
\bottomrule\end{tabular}}\end{center}

\subsection{COCO / DETR}
\begin{center}\adjustbox{max width=\linewidth}{%
\begin{tabular}{@{}ll l rrr@{}}\toprule
Size & Mode & & Padded & Native jagged & DanLing \\ \midrule
official & eager & samples/s & 78.1 & \textsc{ne} & 81.8 \\
 &  & GiB alloc/res. & 7.79 / 14.43 & -- & 1.83 / 4.02 \\
official & compiled & samples/s & \textsc{bl} & \textsc{ne} & \textsc{bl} \\
 &  & GiB alloc/res. & -- & -- & -- \\
\bottomrule\end{tabular}}\end{center}

\subsection{ADE20K / FCN-ResNet}
\begin{center}\adjustbox{max width=\linewidth}{%
\begin{tabular}{@{}ll l rrr@{}}\toprule
Size & Mode & & Padded & Native jagged & DanLing \\ \midrule
18 & eager & samples/s & 135.2 & \textsc{ne} & 297.8 \\
 &  & GiB alloc/res. & 9.89 / 50.46 & -- & 7.06 / 23.40 \\
18 & compiled & samples/s & \textsc{cf} & \textsc{ne} & \textsc{cf} \\
 &  & GiB alloc/res. & -- & -- & -- \\
50 & eager & samples/s & 63.7 & \textsc{ne} & 132.4 \\
 &  & GiB alloc/res. & 28.38 / 63.50 & -- & 12.25 / 26.61 \\
50 & compiled & samples/s & \textsc{nd} & \textsc{ne} & \textsc{cf} \\
 &  & GiB alloc/res. & -- & -- & -- \\
101 & eager & samples/s & 43.4 & \textsc{ne} & 81.2 \\
 &  & GiB alloc/res. & 46.07 / 77.27 & -- & 17.21 / 31.90 \\
101 & compiled & samples/s & \textsc{cf} & \textsc{ne} & \textsc{cf} \\
 &  & GiB alloc/res. & -- & -- & -- \\
152 & eager & samples/s & 33.8 & \textsc{ne} & 59.8 \\
 &  & GiB alloc/res. & 61.67 / 77.51 & -- & 21.60 / 37.92 \\
152 & compiled & samples/s & \textsc{nr} & \textsc{ne} & \textsc{cf} \\
 &  & GiB alloc/res. & -- & -- & -- \\
\bottomrule\end{tabular}}\end{center}

\subsection{Conventional spatial padding}
\label{app:spatial-baseline}

\begin{table}[htbp]
\centering
\caption{\textbf{Matched and conventional spatial padding.} Cells are full-trace mean ms/step / peak allocated GiB.
The conventional workflow changes boundary and interpolation semantics, so it is a different computation rather than a faster version of the same one.}
\adjustbox{max width=\linewidth}{\begin{tabular}{@{}lrrr@{}}
\toprule
Model & Matched padded & Conventional padded & DanLing \\
\midrule
FCN-R50 & 504.1 / 28.38 & 364.8 / 21.34 & 243.8 / 12.25 \\
DETR-R50 & 403.8 / 7.79 & 313.2 / 4.21 & 387.3 / 1.83 \\
\bottomrule
\end{tabular}
}
\end{table}

The conventional FCN workflow applies interpolation over the padded envelope, whereas the matched reference applies it at the valid image extent.
Intermediate convolution and pooling boundaries also differ.
Frozen BatchNorm eliminates one source of coupling but does not identify every source of the observed differences.
DETR comparisons include predictions, matching, loss and gradients.
WMT has no corresponding spatial-boundary comparator.

\subsection{Execution and granularity controls}
\label{app:attribution}

\begin{table}[htbp]
\centering
\caption{\textbf{Five-way execution controls.} Cells are full-trace samples/s.
Random and bucketed policies are distinct frozen batch partitions of the same measured example cohort.}
\adjustbox{max width=\linewidth}{\begin{tabular}{@{}lllrrrrr@{}}
\toprule
Dataset & Batching & Mode & Padded & Attention-only & Explicit packed & DanLing & Native jagged \\
\midrule
IMDB & random & eager & 196.6 & 288.9 & 843.0 & 774.6 & 448.8 \\
IMDB & random & compiled & 208.9 & 316.1 & 899.4 & 901.1 & \textsc{g} \\
IMDB & bucket & eager & 706.3 & 644.3 & 824.9 & 710.0 & 431.2 \\
IMDB & bucket & compiled & 729.9 & 683.0 & 851.7 & 850.5 & \textsc{g} \\
WMT14 & random & eager & 1317.4 & 1139.4 & 1694.9 & 898.8 & 341.8 \\
WMT14 & random & compiled & 39.3 & 136.0 & 2369.3 & 2404.4 & 2352.3 \\
WMT14 & bucket & eager & 1497.6 & 1216.6 & 1666.5 & 882.8 & 311.9 \\
WMT14 & bucket & compiled & 30.0 & 60.7 & 137.4 & 143.7 & 184.9 \\
\bottomrule
\end{tabular}
}
\end{table}

Attention-only packing includes Q/K/V unpadding and output repadding.
Explicit packing and DanLing keep surrounding feature computations packed.
The attention-only, explicit-packed and DanLing controls request native variable-length attention; padded attention runs masked scaled dot-product attention over the padded envelope, and native jagged follows its own SDPA dispatch.
Compiled IMDB reaches similar DanLing and explicit-packed throughput in both policies.
The IMDB random-policy cells repeat the BERT-Base configuration of Figure~\ref{fig:headline} in a separate run, so their ratios differ slightly from that figure's.
WMT combines a larger eager implementation gap with timed compilation outliers in several compiled methods.
Method-level kernel and metadata differences remain part of these comparisons.

The matched-granularity control supports two further comparisons on the Pairformer workload: DanLing with per-sample calls against explicit packing with per-sample calls isolates the interface, and DanLing with segmented calls against DanLing with per-sample calls isolates the execution strategy.
Table~\ref{tab:attribution} reports the three step times for each square regime.

\begin{table}[htbp]
\centering
\caption{\textbf{Matched-granularity control on the four-block Pairformer-style workload, eager execution.} Full-trace mean ms/step.
Explicit packing and DanLing per-sample call each OOps triangle operator once per sample; DanLing segmented calls it once per batch.
Interface cost is the per-sample DanLing slowdown relative to explicit packing; granularity gain is per-sample over segmented DanLing time.
The three methods share one run per regime, separate from Table~\ref{tab:pairformer}, so both ratios come from that run and its times differ slightly from that table's. Ratios use unrounded times.}
\label{tab:attribution}
\adjustbox{max width=\linewidth}{\begin{tabular}{@{}lrrrrr@{}}
\toprule
Shape regime & Explicit packed & DanLing per-sample & DanLing segmented & Interface cost & Granularity gain \\
\midrule
Square, near-uniform ($B=8$) & 404.8 & 406.8 & 265.4 & 0.49\% & 1.53$\times$ \\
Square, moderate ($B=8$) & 396.4 & 399.9 & 210.3 & 0.86\% & 1.90$\times$ \\
Square, high variation ($B=8$) & 391.7 & 398.0 & 148.3 & 1.60\% & 2.68$\times$ \\
Square, extreme variation ($B=16$) & 801.3 & 804.8 & 204.1 & 0.44\% & 3.94$\times$ \\
\bottomrule
\end{tabular}
}
\end{table}

\end{document}